\def\year{2020}\relax
\documentclass[letterpaper]{article} 
\usepackage{aaai20}  
\usepackage{times}  
\usepackage{helvet} 
\usepackage{courier}  
\usepackage[hyphens]{url}  
\usepackage{graphicx} 
\usepackage{graphicx}  
\usepackage{times}
\usepackage{latexsym}
\usepackage{graphicx}
\usepackage{booktabs}
\usepackage{multirow}
\usepackage{mathrsfs}
\usepackage{amsmath}
\usepackage{hyperref}
\usepackage{amssymb}
\usepackage{balance}
\usepackage{url}
\usepackage{bm}
\usepackage{tabularx}
\usepackage{float}
\usepackage{amsmath}
\usepackage{amsthm}
\usepackage{amsfonts}
\usepackage{amssymb}
\usepackage{graphicx}
\usepackage{booktabs}
\usepackage{placeins}
\usepackage{multirow}
\usepackage[ruled, linesnumbered]{algorithm2e}
\usepackage{epstopdf}
\usepackage{color}
\usepackage{subcaption}
\usepackage[T1]{fontenc}
\usepackage{amsmath,amsfonts,bm}
\usepackage{balance}
\usepackage{enumitem}
\usepackage{xcolor}
\usepackage{balance}
\usepackage{subcaption}
\usepackage{enumitem}
\usepackage{graphics}
\usepackage{url}
\usepackage{makecell}
\usepackage[utf8]{inputenc}
\usepackage[english]{babel}

\title{InteractE: Improving Convolution-based Knowledge Graph Embeddings \\ by Increasing Feature Interactions}

\begin{document}
\newcommand{\refalg}[1]{Algorithm \ref{#1}}
\newcommand{\refeqn}[1]{Equation \ref{#1}}
\newcommand{\reffig}[1]{Figure \ref{#1}}
\newcommand{\reftbl}[1]{Table \ref{#1}}
\newcommand{\refsec}[1]{Section \ref{#1}}


\newcommand{\datafb}{FB15k}
\newcommand{\datawn}{WN18}
\newcommand{\datafbn}{FB15k-237}
\newcommand{\datawnn}{WN18RR}
\newcommand{\datayago}{YAGO3-10}

\newcommand{\citet}[1]{\cite{#1}}

\newcommand{\emb}[1]{\ensuremath{\bm{e}_{#1}}}
\newcommand{\remb}[1]{\ensuremath{\bm{#1}_{r}}}

\newcommand{\reminder}[1]{\textcolor{red}{[[ #1 ]]}\typeout{#1}}
\newcommand{\reminderR}[1]{\textcolor{gray}{[[ #1 ]]}\typeout{#1}}

\newcommand{\add}[1]{\textcolor{red}{#1}\typeout{#1}}
\newcommand{\remove}[1]{\sout{#1}\typeout{#1}}

\newcommand{\m}[1]{\mathcal{#1}}
\newcommand{\method}{InteractE}

\newcommand{\problem}{DD}
\newcommand{\problemfull}{Document Dating}

\newtheorem{theorem}{Theorem}[section]
\newtheorem{claim}[theorem]{Claim}

\newcommand{\tensor}{\mathcal{X}}
\newcommand{\Real}{\ensuremath{\mathbb{R}}}
\newcommand{\Natural}{\ensuremath{\mathbb{N}}}
\newcommand{\Complex}{\ensuremath{\mathbb{C}}}
\newcommand{\tdot}[3]{\ensuremath{\langle #1, #2, #3 \rangle}}
\newcommand{\doubledot}[2]{\ensuremath{\langle #1, #2 \rangle}}
\newcommand{\bigO}[1]{\mathcal{O}(#1)}

\newcommand{\tuples}{\mathbb{T}}

\newcommand{\argmax}{arg\,max}
\newcommand\norm[1]{\left\lVert#1\right\rVert}
\newcommand{\note}[1]{\textcolor{blue}{#1}}

\newcommand*{\Scale}[2][4]{\scalebox{#1}{$#2$}}%
\newcommand*{\Resize}[2]{\resizebox{#1}{!}{$#2$}}%

\makeatletter
\newcommand{\ostar}{\mathbin{\mathpalette\make@circled\star}}
\newcommand{\oast}{\mathbin{\mathpalette\make@circled\ast}}
\newcommand{\make@circled}[2]{%
  \ooalign{$\m@th#1\smallbigcirc{#1}$\cr\hidewidth$\m@th#1#2$\hidewidth\cr}%
}
\newcommand{\smallbigcirc}[1]{%
  \vcenter{\hbox{\scalebox{0.77778}{$\m@th#1\bigcirc$}}}%
}
\makeatother

\def\mat#1{\mbox{\bf #1}}

\theoremstyle{definition}
\newtheorem{definition}{Definition}[section]
 
\theoremstyle{proposition}
\newtheorem{proposition}{Proposition}[section]
\newtheorem*{lemma*}{Lemma}

\theoremstyle{remark}
\newtheorem*{remark}{Remark}
\newtheorem{case}{Case}
\newtheorem{subcase}{Case}
\numberwithin{subcase}{case}

\author{Shikhar Vashishth$^{1}$\thanks{ contributed equally to this paper.} \quad Soumya Sanyal$^{1*}$ \quad Vikram Nitin$^{2}$ \vspace{1mm}\\  \Large{\textbf{Nilesh Agrawal}$^1$ \quad \textbf{Partha Talukdar}}$^1$\\
	$^1$ Indian Institute of Science, 
	$^2$ Columbia University \\
	{\tt \small \{shikhar,soumyasanyal,anilesh,ppt\}@iisc.ac.in} \\
	{\tt \small vikram.nitin@columbia.edu}
} 

\maketitle

\begin{abstract}
Most existing knowledge graphs suffer from incompleteness, which can be alleviated by inferring missing links based on known facts. One popular way to accomplish this is to generate low-dimensional embeddings of entities and relations, and use these to make inferences. ConvE, a recently proposed approach, applies convolutional filters on 2D reshapings of entity and relation embeddings in order to capture rich interactions between their components. However, the number of interactions that ConvE can capture is limited. In this paper, we analyze how increasing the number of these interactions affects link prediction performance, and utilize our observations to propose \method{}. \method{} is based on three key ideas -- feature permutation, a novel feature reshaping, and circular convolution. Through extensive experiments, we find that \method{} outperforms state-of-the-art convolutional link prediction baselines on \datafbn{}. Further, \method{} achieves an MRR score that is $9$\%, $7.5$\%, and $23$\% better than ConvE on the \datafbn{}, \datawnn{} and \datayago{} datasets respectively. The results validate our central hypothesis -- that increasing feature interaction is beneficial to link prediction performance. We make the source code of \method{} available to encourage reproducible research.
\end{abstract}

\section{Introduction}
\label{sec:introduction}

Knowledge graphs (KGs) are structured representations of facts, where nodes represent entities and edges represent relationships between them. This can be represented as a collection of triples \((s, r, o)\), each representing a relation \(r\) between a "subject-entity" \(s\) and an "object-entity" \(o\). Some real-world knowledge graphs include Freebase \cite{freebase}, WordNet \cite{wordnet}, YAGO \cite{yago}, and NELL \cite{nell}. KGs find application in a variety of tasks, such as relation extraction \cite{distant_supervision2009}, question answering \cite{qa_kg_1,qa_kg_2}, recommender systems \cite{kb-recommender} and dialog systems \cite{kg_in_dialog}.

However, most existing KGs are incomplete \cite{kg_incomp1}. The task of \textit{link prediction} alleviates this drawback by inferring missing facts based on the known facts in a KG. A popular approach for solving this problem involves learning a low-dimensional representation for all entities and relations and utilizing them to predict new facts. In general, most existing link prediction methods learn to embed KGs by optimizing a score function which assigns higher scores to true facts than invalid ones. These score functions can be classified as \textit{translation distance based} \cite{transe,transg,transh} or \textit{semantic matching  based} \cite{hole,analogy}.

\begin{figure*}[t]
	\centering
	\includegraphics[width=\textwidth]{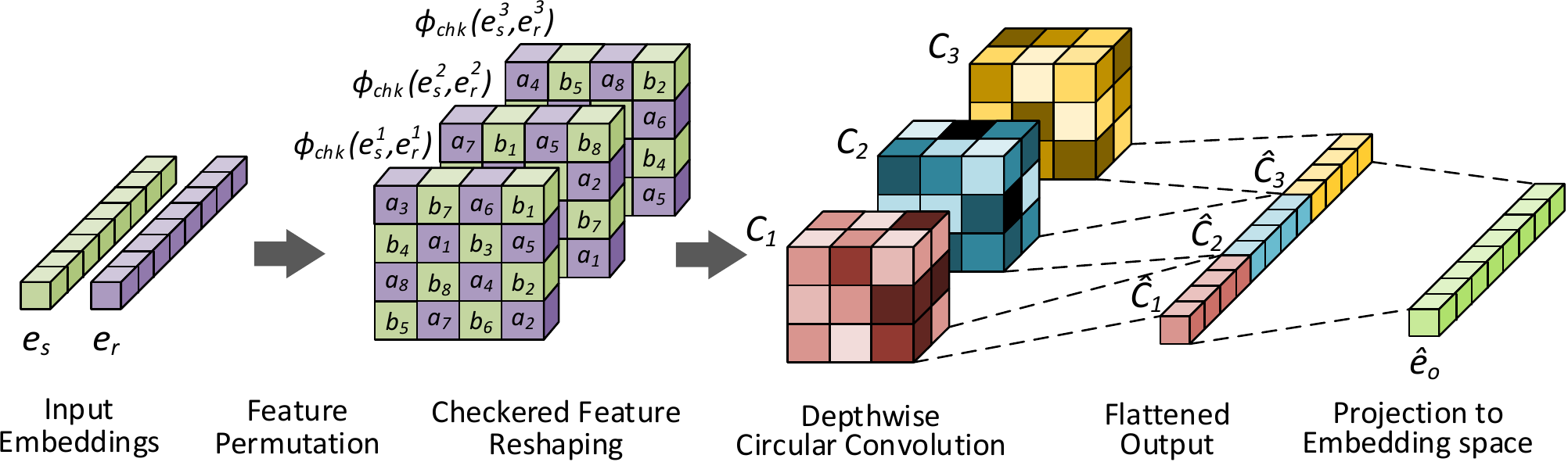}
	\caption{\label{fig:model_overview} Overview of \method{}. Given entity and relation embeddings ($\bm{e}_s$ and $\bm{e}_r$ respectively), \method{} generates multiple permutations of these embeddings and reshapes them using a "Checkered" reshaping function ($\phi_{chk}$). Depth-wise circular convolution is employed to convolve each of the reshaped permutations ($C_i$), which are then flattened ($\hat{C_i}$) and fed to a fully-connected layer to generate the predicted object embedding ($\widehat{\emb{o}}$). Please refer to \refsec{sec:overview} for details.}
\end{figure*}

Recently, neural networks have also been utilized to learn the score function \cite{neural_tensor_network,chandrahas2017,conve}. The motivation behind these approaches is that shallow methods like TransE \cite{transe} and DistMult \cite{distmult} are limited in their expressiveness. As noted in \citet{conve}, the only way to remedy this is to increase the size of their embeddings, which leads to an enormous increase in the number of parameters and hence limits their scalability to larger knowledge graphs.

Convolutional Neural Networks (CNN) have the advantage of using multiple layers, thus increasing their expressive power, while at the same time remaining parameter-efficient. \citet{conve} exploit these properties and propose ConvE - a model which applies convolutional filters on stacked 2D reshapings of entity and relation embeddings. Through this, they aim to increase the number of interactions between components of these embeddings.

In this paper, we conclusively establish that increasing the number of such interactions is beneficial to link prediction performance, and show that the number of interactions that ConvE can capture is limited. We propose \method{}, a novel CNN based KG embedding approach which aims to further increase the interaction between relation and entity embeddings.
Our contributions are summarized as follows: 

\begin{enumerate}[itemsep=2pt,parsep=0pt,partopsep=0pt,leftmargin=*,topsep=0.2pt]
	\item We propose \method{}, a method that augments the expressive power of ConvE through three key ideas -- feature permutation, "checkered" feature reshaping, and circular convolution.
	\item We provide a precise definition of an interaction, and theoretically analyze \method{} to show that it increases interactions compared to ConvE. Further, we establish a correlation between the number of \textit{heterogeneous interactions} (refer to Def. \ref{def:interactions}) and link prediction performance.
	\item Through extensive evaluation on various link prediction datasets, we demonstrate \method{}'s effectiveness (Section \ref{sec:results}).
\end{enumerate}
$ $\\
We have made available the source code of \method{} and datasets used in the paper as a supplementary material.

\section{Related Work}
\label{sec:related_work}


\noindent \textbf{Non-Neural:} Starting with TransE \cite{transe}, there have been multiple proposed approaches that use simple operations like dot products and matrix multiplications to compute a score function. Most approaches embed entities as vectors, whereas for relations, vector \cite{transe,hole}, matrix \cite{distmult,analogy} and tensor \cite{rel_as_tensor} representations have been explored. For modeling uncertainty of learned representations, Gaussian distributions \cite{gaussian_kg,transg} have also been utilized.
Methods like TransE \cite{transe} and TransH \cite{transh} utilize a translational objective for their score function, while DistMult \cite{distmult} and ComplEx \cite{complex} use a bilinear diagonal based model.

\noindent \textbf{Neural Network based:} Recently, Neural Network (NN) based score functions have also been proposed. Neural Tensor Network \cite{neural_tensor_network} combines entity and relation embeddings by a relation-specific tensor which is given as input to a non-linear hidden layer for computing the score. \citet{kg_incomp1,chandrahas2017} also utilize a Multi-Layer Perceptron for modeling the score function. 


\noindent \textbf{Convolution based:} Convolutional Neural Networks (CNN) have also been employed for embedding Knowledge Graphs. ConvE \cite{conve} uses convolutional filters over reshaped subject and relation embeddings to compute an output vector and compares this with all other entities in the knowledge graph.
\citeauthor{sacn_paper} propose ConvTransE a variant of the ConvE score function. They eschew 2D reshaping in favor of directly applying convolution on the stacked subject and relation embeddings. Further, they propose SACN which utilizes \textit{weighted graph convolution} along with ConvTransE.


ConvKB \cite{convkb} is another convolution based method which 
applies convolutional filters of width 1 on the stacked subject, relation and object embeddings for computing score. As noted in \cite{sacn_paper}, although ConvKB was claimed to be superior to ConvE, its performance is not consistent across different datasets and metrics. Further, there have been concerns raised about the validity of its evaluation procedure\footnote{\url{https://openreview.net/forum?id=HkgEQnRqYQ&noteId=HklyVUAX2m}} Hence, we do not compare against it in this paper.
A survey of all variants of existing KG embedding techniques can be found in \cite{survey2016nickel,survey2017}.


\section{Background}
\label{sec:background}
\noindent \textbf{KG Link Prediction:} Given a Knowledge Graph (KG) $\m{G} = (\m{E}, \m{R, \m{T} })$, where $\m{E}$ and $\m{R}$ denote the set of entities and relations, and $\m{T}$ denotes the triples (facts) of the form $\{(s,r,o)\} \subset \m{E}\times \m{R} \times \m{E}$, the task of \textit{link prediction} is to predict new facts $(s',r',o')$ such that $s',o' \in \m{E}$ and $r' \in \m{R}$, based on the existing facts in KG. Formally, the task can be modeled as a ranking problem, where the goal is to learn a function $\psi(s,r,o): \m{E}\times \m{R} \times \m{E} \rightarrow \mathbb{R}$ which assigns higher scores to true or likely facts  than invalid ones.

\begin{table}[t]
	\begin{center}
		\begin{tabular}{lc}
			
			\toprule
			\multicolumn{1}{c}
			{ Model} &{Scoring Function} $\psi(\emb{s}, \emb{r}, \emb{o})$												\\
			\toprule
			TransE 		  &$\norm{\emb{s} + \emb{r} - \emb{o}}_{p}$																\\
			DistMult 	  &$\tdot{\emb{s}}{\emb{r}}{\emb{o}}$						 												\\
			HolE 		   &$\doubledot{\emb{r}}{\emb{s} \ast \emb{o}}$			 												\\
			ComplEx 	&$\mathrm{Re}(\tdot{\emb{s}}{\emb{r}}{\emb{o}})$		 												\\
			ConvE 		 &$f (\mathrm{vec} ( f ([ \overline{\emb{s}} ; \overline{\emb{r}} ] \star w )) \mathbf{W} )\emb{o}$	\\
			RotatE 		  &$-\norm{\emb{s}  \circ \emb{r} - \emb{o}}^2$	\\
			\midrule
			\method{} 	&$g(\mathrm{vec}(f(\phi(\bm{\m{P}}_k) \ostar w))\bm{W})\bm{e}_o$				\\
			\bottomrule
		\end{tabular}
		\caption{\label{tab:score_functions} The scoring functions $\psi(s, r, o)$ of various knowledge graph embedding methods. Here, $\emb{s}, \emb{r}, \emb{o} \in \mathbb{R}^d$ except for ComplEx and RotatE, where they are complex vectors $(\mathbb{C}^d)$, $\ast$ denotes circular-correlation, $\star$ denotes convolution, $\circ$ represents Hadamard product and $\star$ denotes depth-wise circular convolution operation.		}
	\end{center}
\end{table}

Most existing KG embedding approaches define an encoding for all entities and relations, i.e., $\bm{e}_{s}, \bm{e}_{r} \ \forall s \in \m{E}, r \in \m{R}$. Then, a score function $\psi(s,r,o)$ is defined to measure the validity of triples. Table \ref{tab:score_functions} lists some of the commonly used score functions. Finally, to learn the entity and relation representations, an optimization problem is solved for maximizing the plausibility of the triples $\m{T}$ in the KG.

\noindent \textbf{ConvE}: In this paper, we build upon ConvE \cite{conve}, which models interaction between entities and relations using 2D Convolutional Neural Networks (CNN). The score function used is defined as follows:
\[
\psi(s, r, o) = f(\mathrm{vec}(f([\overline{\bm{e}_s};\overline{\bm{e}_r} \star w]))\bm{W})\bm{e_o},
\]
where, $\overline{\bm{e_s}} \in \mathbb{R}^{d_w \times d_h}$, $\overline{\bm{e_r}} \in \mathbb{R}^{d_w \times d_h}$ denote 2D reshapings of $\bm{e_s} \in \mathbb{R}^{d_{w}d_{h} \times 1}$, $\bm{e_r} \in \mathbb{R}^{d_{w}d_{h} \times 1}$, and $(\star
)$ denotes the convolution operation. The 2D reshaping enhances the interaction between entity and relation embeddings which has been found to be helpful for learning better representations \cite{hole}. 

\section{Notation and Definitions}
\label{sec:notations}

Let $\emb{s} = (a_1, ... , a_d), \emb{r} = (b_1, ... , b_d)$, where $a_i, b_i \in \mathbb{R} \ \forall i$, be an entity and a relation embedding respectively, and let $\bm{w} \in \mathbb{R}^{k \times k}$ be a convolutional kernel of size $k$.
Further, we define that a matrix $M_k \in \mathbb{R}^{k \times k}$ is a $k$-submatrix of another matrix $N \in \mathbb{R}^{m \times n}$  if $\exists \ i,j$ such that $M_k = N_{i:i+k,\ j:j+k}$. We denote this by $M_k \subseteq N$.

\begin{figure}[t]
	\centering
	\includegraphics[width=\columnwidth]{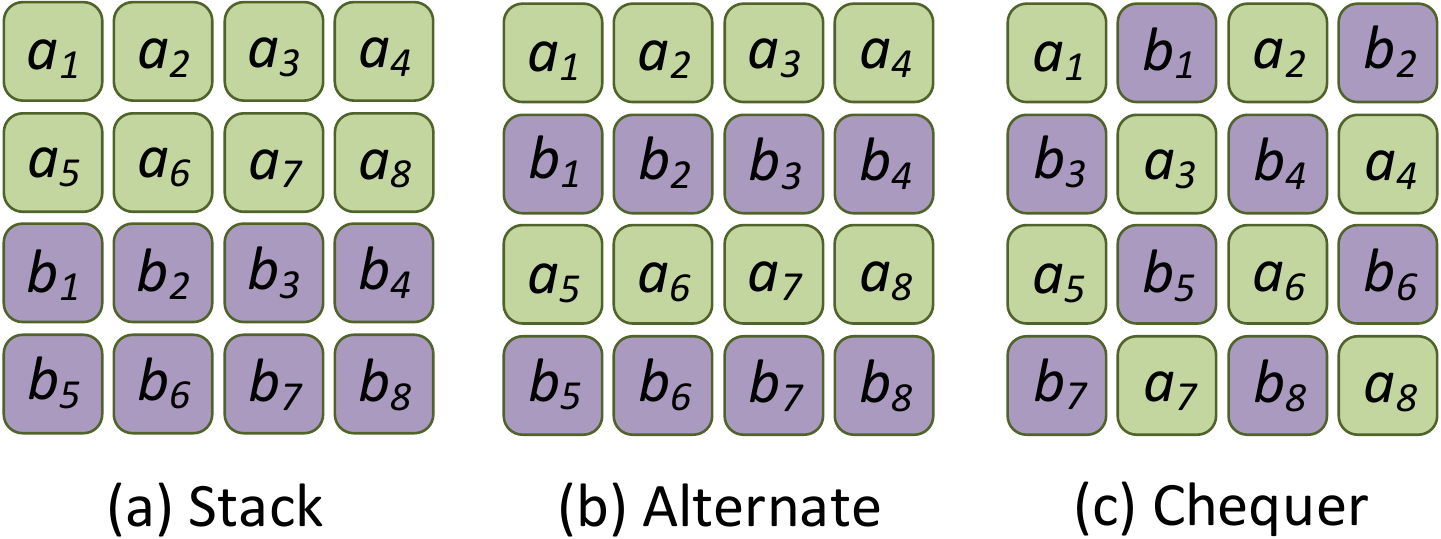}
	\caption{\label{fig:reshaping}Different types of reshaping functions we analyze in this paper. Here, $\emb{s} =  (a_1, ... , a_8),  \emb{r} = (b_1, ... , b_8)$, and $m = n = 4$. Please refer to Section \ref{sec:notations}	for more details.}
\end{figure}

\theoremstyle{definition}
\begin{definition}{\textbf{(Reshaping Function)}}
	\label{def:reshape_func}
	A reshaping function $\phi: \mathbb{R}^d \times \mathbb{R}^d \to \mathbb{R}^{m\times n}$ transforms embeddings $\emb{s}$ and $\emb{r}$ into a matrix $\phi(\emb{s}, \emb{r})$, where $m \times n = 2d$. For conciseness, we abuse notation and represent $\phi(\emb{s}, \emb{r})$ by $\phi$. We define three types of reshaping functions.

	\begin{itemize}[itemsep=3pt,parsep=3pt,partopsep=3pt,leftmargin=10pt,topsep=3pt]
		\item \textbf{Stack} ($\phi_{stk}$) reshapes each of $\emb{s}$ and $\emb{r}$ into a matrix of shape $(m/2) \times n$, and stacks them along their height to yield an $m \times n$ matrix (Fig. \ref{fig:reshaping}a). This is the reshaping function used in \citet{conve}.
		\item \textbf{Alternate} ($\phi_{alt}^{\tau}$) reshapes $\emb{s}$ and $\emb{r}$ into matrices of shape $(m/2) \times n$, and stacks $\tau$ rows of $\emb{s}$ and $\emb{r}$ alternately. In other words, as we decrease $\tau$, the "frequency" with which rows of $\emb{s}$ and $\emb{r}$ alternate increases. We denote $\phi_{alt}^1 (\tau =1)$ as $\phi_{alt}$ for brevity (Fig. \ref{fig:reshaping}b).
		\item \textbf{Chequer} ($\phi_{chk}$) arranges $\emb{s}$ and $\emb{r}$ such that no two adjacent cells are occupied by components of the same embedding (Fig. \ref{fig:reshaping}c).
	\end{itemize}
\end{definition}

\begin{definition}{\textbf{(Interaction)}}
	\label{def:interactions}
	An interaction is defined as a triple $(x, y, M_k)$, such that $M_k\subseteq \phi(\emb{s}, \emb{r})$ is a $k$-submatrix of the reshaped input embeddings; $x,y \in M_k$ and are distinct components of $\emb{s}$ or $\emb{r}$.
	The number of interactions $\mathcal{N}(\phi, k)$ is defined as the cardinality of the set of all possible triples. Note that $\phi$ can be replaced with $\Omega(\phi)$ for some padding function $\Omega$.

	An interaction $(x, y, M_k)$ is called \textbf{heterogeneous} if $x$ and $y$ are components of $\emb{s}$ and $\emb{r}$ respectively, or vice-versa. Otherwise, it is called \textbf{homogeneous}. We denote the number of heterogeneous and homogeneous interactions as  $\mathcal{N}_{het}(\phi, k)$ and $\mathcal{N}_{homo}(\phi, k)$ respectively. For example, in a $3 \times 3$ matrix $M_3$, if there are $5$ components of $\emb{s}$ and $4$ of $\emb{r}$, then the number of heterogeneous and homogeneous interactions are: $\m{N}_{het} = 2(5 \times 4) = 40$, and $\m{N}_{homo} = 2\left[\binom{5}{2} + \binom{4}{2}\right] = 32$. Please note that the sum of total number of heterogenous and homogenous interactions in a reshaping function is constant and is equal to $2 \binom{k^2}{2}$, i.e., $\mathcal{N}_{het}(\phi, k) + \mathcal{N}_{homo}(\phi, k) = 2 \binom{k^2}{2}$.
	
\end{definition}

\section{\method{} Overview}
\label{sec:overview}

Recent methods \cite{distmult,hole} have demonstrated that expressiveness of a model can be enhanced by increasing the possible interactions between embeddings. ConvE \cite{conve} also exploits the same principle albeit in a limited way, using convolution on 2D reshaped embeddings.
\method{} extends this notion of capturing entity and relation feature interactions using the following three ideas:

\begin{itemize}[itemsep=3pt,parsep=3pt,partopsep=3pt,leftmargin=10pt,topsep=3pt]
	\item \textbf{Feature Permutation:} Instead of using one fixed order of the input, we utilize multiple permutations to capture more possible interactions. 
	\item \textbf{Checkered Reshaping:} We substitute simple feature reshaping of ConvE with checked reshaping and prove its superiority over other possibilities.
	\item \textbf{Circular Convolution:} Compared to the standard convolution, circular convolution allows to capture more feature interactions as depicted in Figure \ref{fig:circ_conv}. The convolution is performed in a depth-wise manner \cite{depthwise_convolution} on different input permutations.
\end{itemize}
\section{\method{} Details}
\label{sec:details}

In this section, we provide a detailed description of the various components of \method{}. The overall architecture is depicted in Fig. \ref{fig:model_overview}. \method{} learns a $d$-dimensional vector representation $(\emb{s}, \emb{r} \in \mathbb{R}^{d})$ for each entity and relation in the knowledge graph, where $d=d_wd_h$.

\subsection{Feature Permutation}
\label{sec:rearrange}
To capture a variety of heterogeneous interactions, \method{} first generates $t$-random permutations of both $e_s$ and $e_r$, denoted by $\bm{\m{P}}_t = [(\emb{s}^1, \emb{r}^1); ... ;(\emb{s}^t, \emb{r}^t) ]$.
Note that with high probability, the sets of interactions within $\phi(\emb{s}^i, \emb{r}^i)$ for different $i$ are disjoint. This is evident because the number of distinct interactions across all possible permutations is very large. So, for $t$ different permutations, we can expect the total number of interactions to be approximately $t$ times the number of interactions for one permutation.

\subsection{Checkered Reshaping}
\label{sec:checkered}
Next, we apply the reshaping operation $\phi_{chk}(\emb{s}^i, \emb{r}^i), \forall i \in \{1, ..., t\}$, and define $\phi(\bm{\m{P}}_t )= [\phi(\emb{s}^1, \emb{r}^1); ... ; \phi(\emb{s}^t, \emb{r}^t)]$. ConvE \cite{conve} uses $\phi_{stk}(\cdot)$ as a reshaping function which has limited interaction capturing ability. On the basis of Proposition \ref{thm:max_best}, we choose to utilize $\phi_{chk}(\cdot)$ as the reshaping function in \method{}, which captures maximum \textit{heterogeneous} interactions between entity and relation features.

\subsection{Circular Convolution}
\label{sec:cconv}
\begin{figure}[t]
	\centering
	\includegraphics[width=\columnwidth]{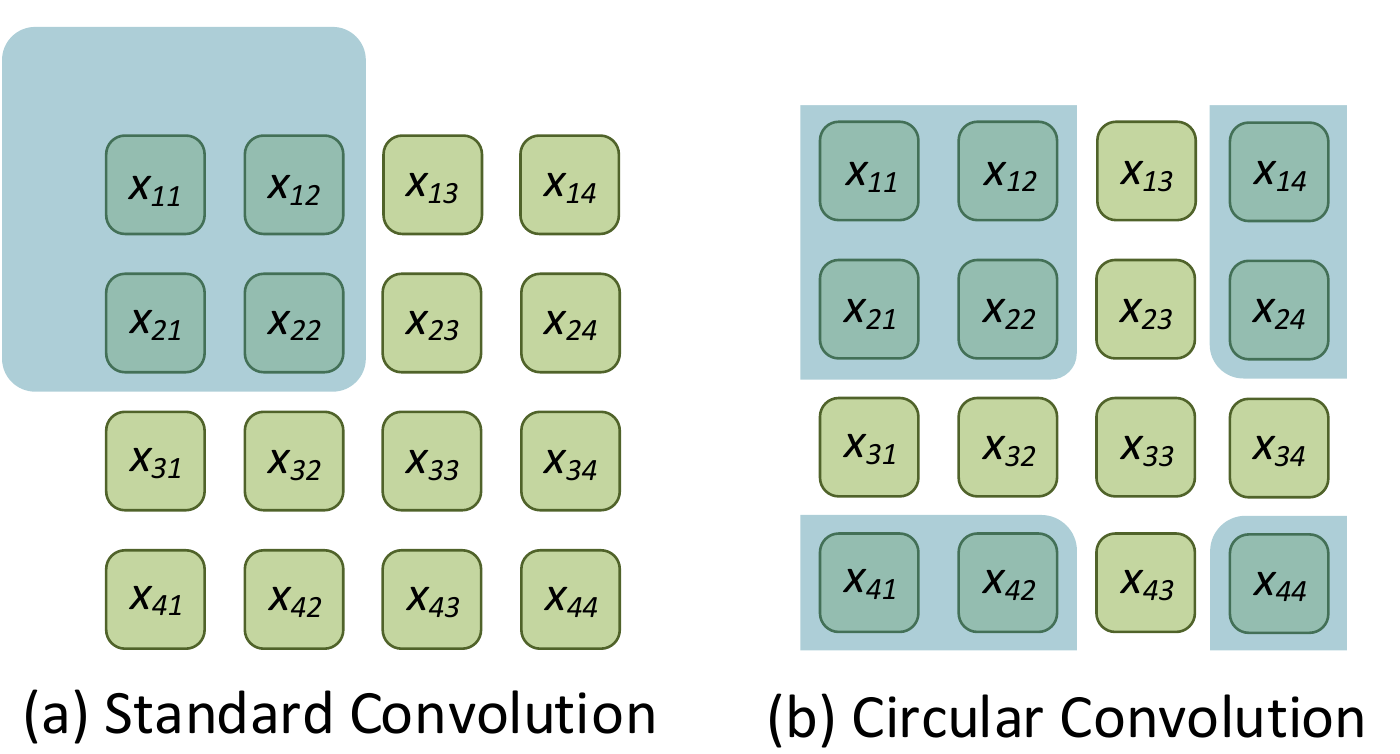}
	\caption{\label{fig:circ_conv} Circular convolution induces more interactions than standard convolution. Here, $X$ is a $4 \times 4$ input matrix with components $x_{ij}$. The shaded region depicts where the filter is applied. Please refer to Section \ref{sec:cconv} for more details.
	}
\end{figure}

Motivated by our analysis in Proposition \ref{thm:circ_comp}, \method{} uses \textit{circular convolution}, which further increases interactions compared to the standard convolution. This has been successfully applied for tasks like image recognition \cite{omnidirectionalwang2018}. Circular convolution on a $2$-dimensional input $\bm{I} \in \mathbb{R}^{m \times n}$ with a filter $w \in \mathbb{R}^{k \times k}$ is defined as:
\[
[\bm{I} \star \bm{w}]_{p,q} = \sum_{i = - \lfloor k/2 \rfloor}^{\lfloor k/2\rfloor} \sum_{j = - \lfloor k/2 \rfloor}^{\lfloor k/2\rfloor}\bm{I}_{[p-i]_m, [q-j]_n} \bm{w}_{i,j},
\]
where, $[x]_{n}$ denotes $x$ modulo $n$ and $\lfloor \cdot \rfloor$ denotes the floor function. Figure \ref{fig:circ_conv} and Proposition \ref{thm:circ_comp} show how circular convolution captures more interactions compared to standard convolution with zero padding.

\method{} stacks each reshaped permutation as a separate channel. For convolving permutations, we apply circular convolution in a \textit{depth-wise} manner \cite{depthwise_convolution}. Although different sets of filters can be applied for each permutation, in practice we find that sharing filters across channels works better as it allows a single set of kernel weights to be trained on more input instances.

\subsection{Score Function}
The output of each circular convolution is flattened and concatenated into a vector. \method{} then projects this vector to the embedding space ($\mathbb{R}^d$). 
Formally, the score function used in \method{} is defined as follows:
\[
\psi(s,r,o) = g(\mathrm{vec}(f(\phi(\bm{\m{P}}_k) \ostar \bm{w}))\bm{W})\bm{e}_o,
\]
where $\ostar$ denotes depth-wise circular convolution, $\mathrm{vec}(\cdot)$ denotes vector concatenation, $\bm{e}_o$ represents the object entity embedding matrix and $W$ is a learnable weight matrix. Functions $f$ and $g$ are chosen to be ReLU and sigmoid respectively. For training, we use the standard binary cross entropy loss with label smoothing.

%
%
%
%

\section{Theoretical Analysis}
\label{sec:theory}

In this section, we analyze multiple variants of 2D reshaping with respect to the number of interactions they induce. We also examine the advantages of using circular padded convolution over the standard convolution. 

For simplicity, we restrict our analysis to the case where the output of the reshaping function is a square matrix, i.e., $m=n$. Note that our results can be extended to the general case as well. Proofs of all propositions herein are included in the supplementary material. 

\begin{proposition}
	\label{thm:cat_alt_comp}
	For any kernel $\bm{w}$ of size $k$, for all $n\ge\left(\frac{5k}{3}-1\right)$ if $k$ is odd and $n\ge\frac{(5k+2)(k-1)}{3k}$ if $k$ is even, the following statement holds:
	\[
	\mathcal{N}_{het}(\phi_{alt}, k) \ge \mathcal{N}_{het}(\phi_{stk}, k)
	\]
\end{proposition}

\begin{proposition}
	\label{thm:alt_tau_comp}
	For any kernel $\bm{w}$ of size $k$ and for all $\tau < \tau'$ ($\tau, \tau' \in \mathbb{N}$), the following statement holds:
	\[
	\mathcal{N}_{het}(\phi_{alt}^{\tau}, k) \ge \mathcal{N}_{het}(\phi_{alt}^{\tau'}, k)
	\]
\end{proposition}

\begin{proposition}
	\label{thm:max_best}
	For any kernel $\bm{w}$ of size $k$ and for all reshaping functions $\phi : \mathbb{R}^d \times \mathbb{R}^d \to \mathbb{R}^{n \times n}$, the following statement holds:
	\[
	\mathcal{N}_{het}(\phi_{chk}, k) \ge \mathcal{N}_{het}(\phi, k)
	\]
\end{proposition}

\begin{proposition}
	\label{thm:circ_comp}
	Let $\Omega_0$, $\Omega_c : \mathbb{R}^{n \times n} \to \mathbb{R}^{(n+p) \times (n+p)}$ denote zero padding and circular padding functions respectively, for some $p > 0$. Then for any reshaping function $\phi$,
	\[
	\mathcal{N}_{het}(\Omega_c(\phi), k) \ge \mathcal{N}_{het}(\Omega_0(\phi), k)
	\]
\end{proposition}

\begin{table*}[t]
	\centering
	\begin{small}
	\resizebox{\textwidth}{!}{
	\begin{tabular}{lcccccccccccc}
		\toprule
		& \multicolumn{4}{c}{\textbf{\datafbn{}}} & \multicolumn{4}{c}{\textbf{\datawnn{}}} & \multicolumn{4}{c}{\textbf{YAGO3-10}} \\ 
		\cmidrule(r){2-5}  \cmidrule(r){6-9} \cmidrule(r){10-13} 
& MRR & MR &H@10 &  H@1 & MRR & MR & H@10  & H@1 & MRR & MR & H@10  & H@1 \\
		\midrule
		DistMult \cite{distmult}	& .241 & 254 & .419 & .155 & .430 & 5110 & .49  & .39  & .34	& 5926	& .54 	& .24 \\
ComplEx	\cite{complex}	& .247 & 339 & .428 & .158 & .44  & 5261 & .51  & .41 & .36	& 6351	& .55	& .26 \\
R-GCN \cite{r_gcn}		& .248 & -   & .417 & .151 & -    & - 	 & -    &  -   & -	& - 	& - 	& - \\
KBGAN \cite{kbgan}		&  .278  & -   & .458  & -  &  .214 & - 	 & .472  & -  & -	& -	& -	& - \\
KBLRN \cite{kblrn}		& .309  & 209   & .493  & .219  &  - & - 	 & -  & -  & -	& -	& -	& - \\
ConvTransE	\cite{sacn_paper} & .33  & -   & .51  & .24  &  .46 & - 	 & .52  & .43  & -	& -	& -	& - \\
SACN \cite{sacn_paper} 		& .35  & -   & .54  & .26  &  .47 & - 	 & .54  & .43  & -	& -	& -	& - \\
RotatE \cite{rotate}		& .338 & 177 & .533 & .241 & \textbf{.476} & \textbf{3340} & \textbf{.571} & .428 & .495	& 1767	& .670	& .402 \\
\midrule 
ConvE \cite{conve}		& .325 & 244 & .501 & .237 &  .43 & 4187 & .52  & .40  & .44	& \textbf{1671}	& .62	& .35 \\
\midrule 
\method{} (Proposed Method)	& \textbf{.354} & \textbf{172} & \textbf{.535} & \textbf{.263} & .463 & 5202 & .528 & \textbf{.430} & \textbf{.541} & 2375 & \textbf{.687} & \textbf{.462}\\
\bottomrule
\addlinespace
\end{tabular}
}
		\caption{\label{tbl:main_result_1} Link prediction results of several models evaluated on \datafbn{}, \datawnn{} and \datayago{}. We find that \method{} outperforms all other methods across metrics on \datafbn{} and in $3$ out of $4$ settings on \datayago{}. Since InteractE generalizes ConvE, we highlight performance comparison between the two methods specifically in the table above. Please refer to Section \ref{sec:main_results} for more details.}
	\end{small}
\end{table*}

\section{Experimental Setup}
\label{sec:experiments}

\subsection{Datasets}
\label{sec:datasets}
In our experiments, following \citet{conve,rotate}, we evaluate on the three most commonly used link prediction datasets. A summary statistics of the datasets is presented in Table \ref{table:datasets}.
\begin{itemize}[itemsep=2pt,parsep=0pt,partopsep=0pt,leftmargin=10pt,topsep=2pt]
	\item \textbf{\datafbn{}} \cite{toutanova} is a improved version of FB15k \cite{transe} dataset where all inverse relations are deleted to prevent direct inference of test triples by reversing training triples.
	\item \textbf{\datawnn{}} \cite{conve} is a subset of WN18 \cite{transe} derived from WordNet \cite{wordnet}, with deleted inverse relations similar to \datafbn{}.
	\item \textbf{YAGO3-10} is a subset of YAGO3 \cite{yago} constitutes entities with at least 10 relations. Triples consist of descriptive attributes of people.
\end{itemize}

\subsection{Evaluation protocol}
\label{sec:evaluation}
Following \cite{transe}, we use the filtered setting, i.e., while evaluating on test triples, we filter out all the valid triples from the candidate set, which is generated by either corrupting the head or tail entity of a triple. The performance is reported on the standard evaluation metrics: Mean Reciprocal Rank (MRR), Mean Rank (MR) and Hits@1, and Hits@10. We report average results across $5$ runs. We note that the variance is substantially low on all the metrics and hence omit it. 

\subsection{Baselines}
\label{sec:baselines}
In our experiments, we compare \method{} against a variety of baselines which can be categorized as:
\begin{itemize}[itemsep=3pt,parsep=0pt,partopsep=0pt,leftmargin=10pt,topsep=2pt]

\item \textbf{Non-neural}: Methods that use simple vector based operations for computing score. For instance, DistMult \cite{distmult}, ComplEx \cite{complex}, KBGAN \cite{kbgan}, KBLRN \cite{kblrn} and RotatE \cite{rotate}.

\item \textbf{Neural}: Methods which leverage a non-linear neural network based architecture in their scoring function. This includes R-GCN \cite{r_gcn}, ConvE \cite{conve}, ConvTransE \cite{sacn_paper}, and SACN \cite{sacn_paper}.

\end{itemize}

\begin{table}[t]
	\centering
	\resizebox{\columnwidth}{!}{
		\begin{tabular}{lrrrrr}
			\toprule
			Dataset 			& \multicolumn{1}{c}{$|\m{E}|$} & \multicolumn{1}{c}{$|\m{R}|$}	&\multicolumn{3}{c}{\# Triples}			\\
			\cmidrule(r){4-6}
			& 	& 		& \multicolumn{1}{c}{Train} & \multicolumn{1}{c}{Valid} & \multicolumn{1}{c}{Test}			\\
			\midrule
			\datafbn{} & 14,541 & 237	& 272,115 & 17,535 & 20,466 \\
			\datawnn{} & 40,943 & 11	& 86,835 & 3,034 & 3,134 \\
			YAGO3-10 & 123,182 & 37	& 1,079,040 & 5,000 & 5,000 \\			
			\bottomrule
		\end{tabular}
	}
	\caption{\label{table:datasets}Details of the datasets used. Please see Section \ref{sec:datasets} for more details.}
\end{table}

\section{Results}
\label{sec:results}

In this section, we attempt to answer the questions below:
\begin{itemize}[itemsep=1pt,topsep=2pt,parsep=0pt,partopsep=0pt,leftmargin=20pt]
	\item[Q1.] How does \method{} perform in comparison to the existing approaches? (Section \ref{sec:main_results})
	\item[Q2.] What is the effect of different feature reshaping and circular convolution on link prediction performance? (Section \ref{sec:ablation_results})
	\item[Q3.] How does the performace of our model vary with number of feature permutations? (Section \ref{sec:perm_kernel_results})
	\item[Q4.] What is the performance of \method{} on different relation types? (Section \ref{sec:results_rel_cat})
\end{itemize}

\begin{figure*}[t!]
	\begin{minipage}{0.49\linewidth}
		\centering
		\includegraphics[width=\linewidth]{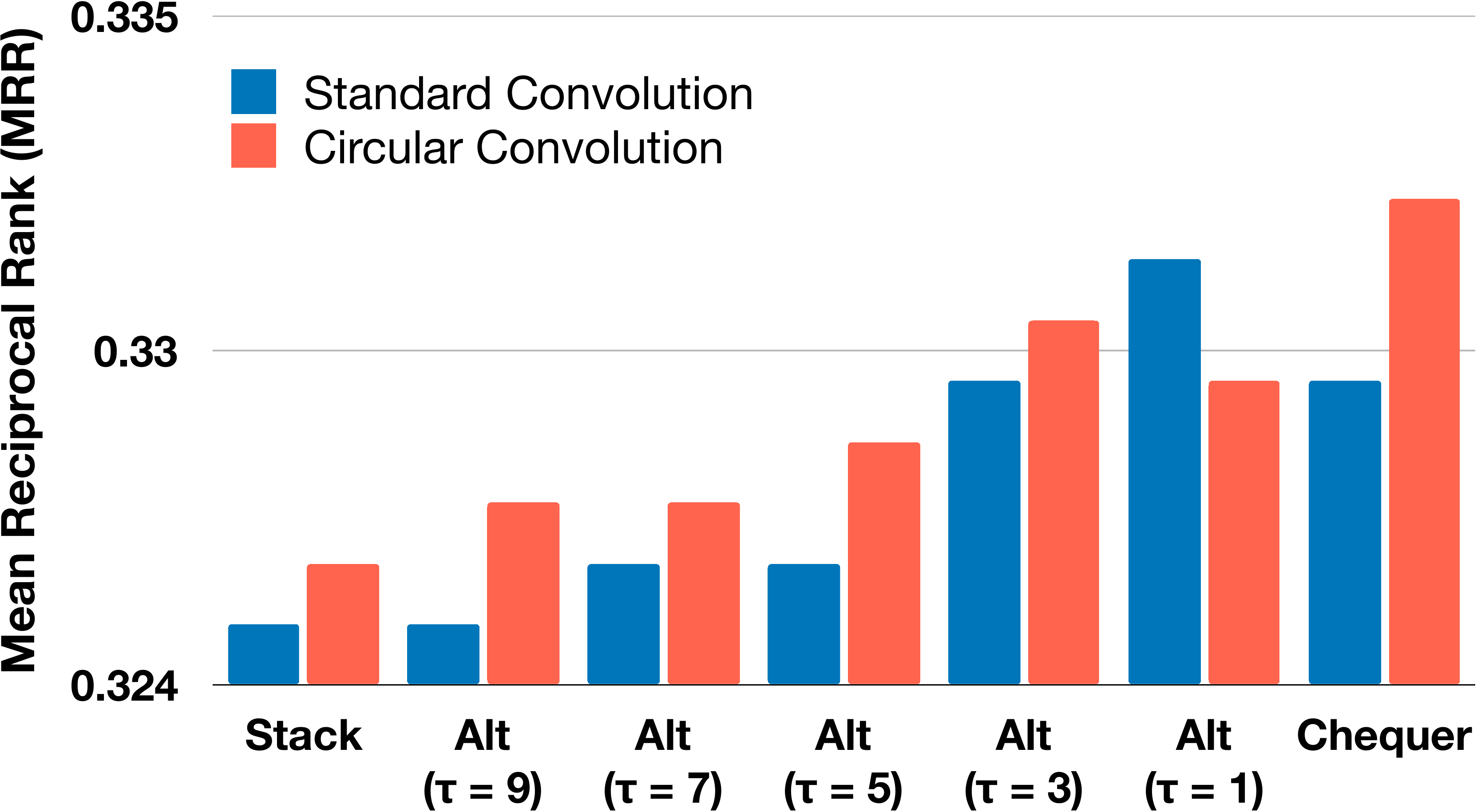}
		\subcaption{\datafbn{} dataset}
	\end{minipage}
	\hfill
	\begin{minipage}{0.49\linewidth}
		\centering
		\includegraphics[width=\linewidth]{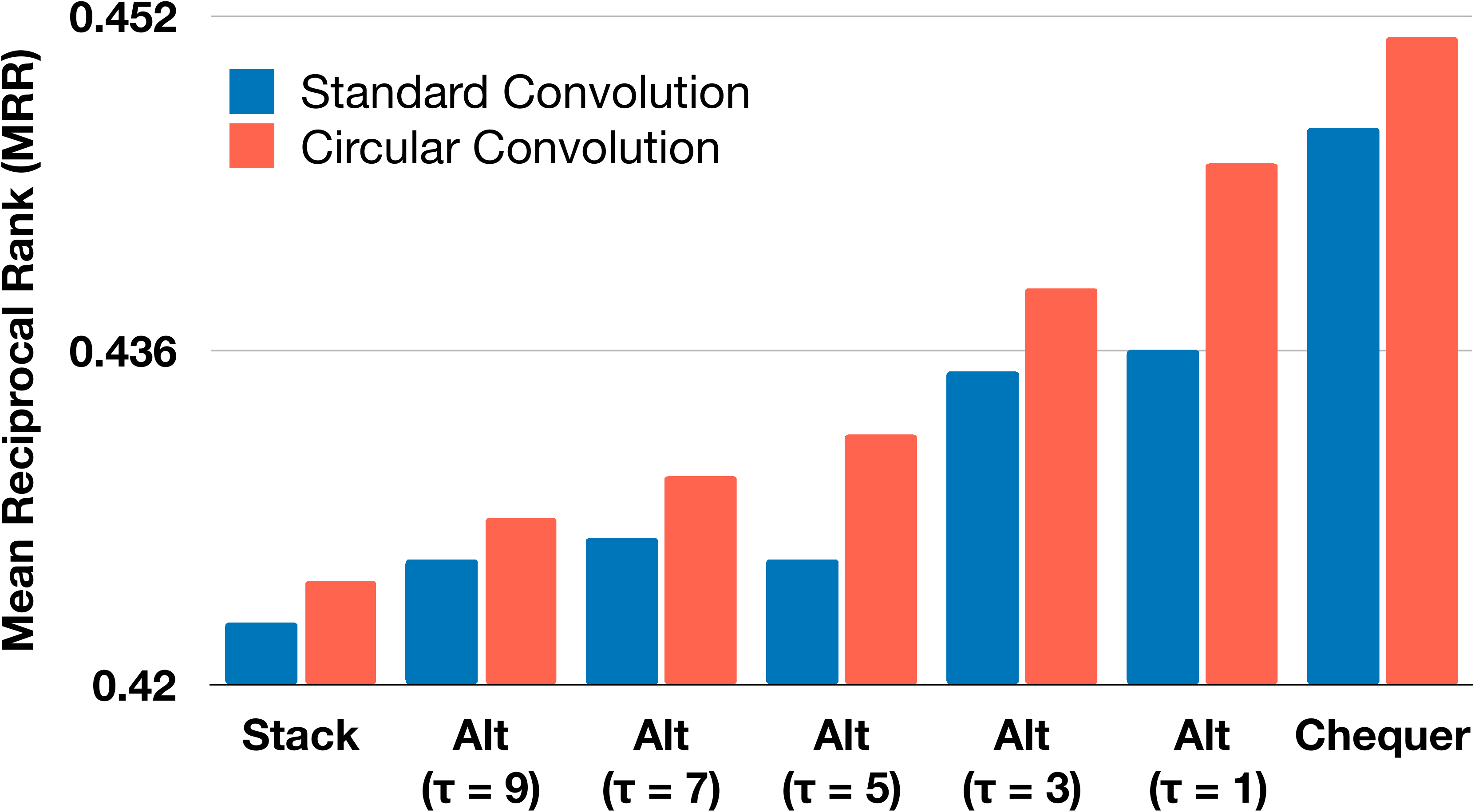}
		\subcaption{\datawnn{} dataset}
	\end{minipage}
	\caption{\label{fig:fact_analysis}Performance with different feature reshaping and convolution operation on validation data of \datafbn{} and \datawnn{}. Stack and Alt denote \textit{Stacked} and \textit{Alternate} reshaping as defined in Section \ref{sec:notations}. As we decrease $\tau$ the number of heterogeneous interactions increases (refer to Proposition \ref{thm:alt_tau_comp}). The results empirically verify our theoretical claim in Section \ref{sec:theory} and validate the central thesis of this paper that increasing \textit{heterogeneous interactions} improves link prediction performance. Please refer to Section \ref{sec:ablation_results} for more details.}
\end{figure*}

\subsection{Performance Comparison}
\label{sec:main_results}

In order to evaluate the effectiveness of \method{}, we compare it against the existing knowledge graph embedding methods listed in Section \ref{sec:baselines}. The results on three standard link prediction datasets are summarized in Table \ref{tbl:main_result_1}. The scores of all the baselines are taken directly from the values reported in the papers \cite{conve,rotate,sacn_paper,kbgan,kblrn}. Since our model builds on ConvE, we specifically compare against it, and find that \method{} outperforms ConvE on all metrics for \datafbn{} and \datawnn{} and on three out of four metrics on \datayago{}. On an average, \method{} obtains an improvement of $9$\%, $7.5$\%, and $23$\% on \datafbn{}, \datawnn{}, and \datayago{} on MRR over ConvE. This validates our hypothesis that increasing heterogeneous interactions help improve performance on link prediction. For \datayago{}, we observe that the MR obtained from \method{} is worse than ConvE although it outperforms ConvE on all other metrics. Simliar trend has been observed in \cite{conve,rotate}.


Compared to other baseline methods, \method{} outperforms them on \datafbn{} across all the metrics and on $3$ out of $4$ metrics on \datayago{} dataset. The below-par performance of \method{} on \datawnn{} can be attributed to the fact that this dataset is more suitable for shallow models as it has very low average relation-specific in-degree. This is consistent with the observations of \citet{conve}.

\subsection{Effect of Feature Reshaping and Circular Convolution}
\label{sec:ablation_results}

In this section, we empirically test the effectiveness of different reshaping techniques we analyzed in Section \ref{sec:theory}. For this, we evaluate different variants of \method{}  on validation data of \datafbn{} and \datawnn{} with the number of feature permutations set to $1$. We omit the analysis on \datayago{} given its large size. The results are summarized in Figure \ref{fig:fact_analysis}. We find that the performance with \textit{Stacked} reshaping is the worst, and it improves when we replace it with \textit{alternate} reshaping. This observation is consistent with our findings in Proposition \ref{thm:cat_alt_comp}. Further, we find that MRR improves on decreasing the value of $\tau$ in alternate reshaping, which empirically validates Proposition \ref{thm:alt_tau_comp}. Finally, we observe that \textit{checkered} reshaping gives the best performance across all reshaping functions for most scenarios, thus justifying Proposition \ref{thm:max_best}.

We also compare the impact of using circular and standard convolution on link prediction performance. The MRR scores are reported in Figure \ref{fig:fact_analysis}. The results show that circular convolution is consistently better than the standard convolution. This also verifies our statement in Proposition \ref{thm:circ_comp}. Overall, we find that increasing interaction helps improve performance on the link prediction task, thus validating the central thesis of our paper.

\begin{figure}[t]
	\centering
	\includegraphics[width=\columnwidth]{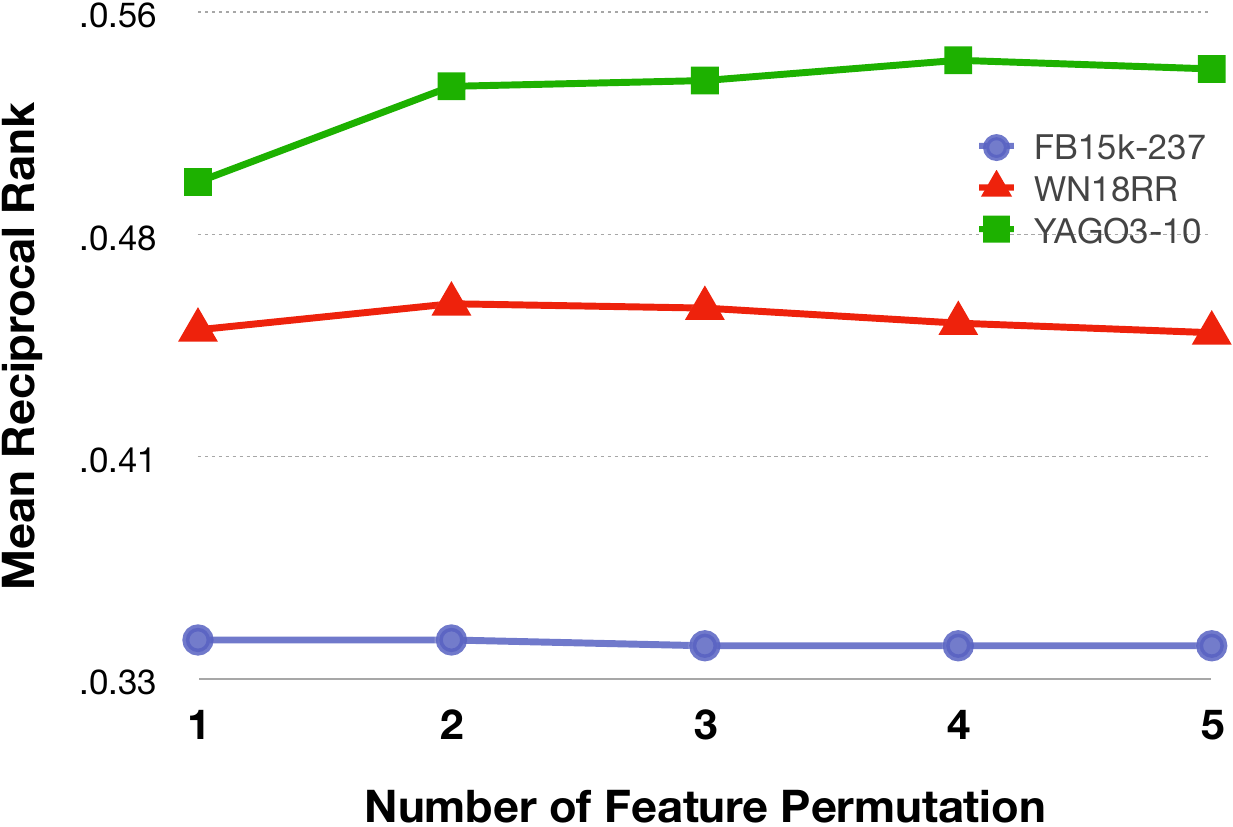}
	\caption{\label{fig:perm_analysis}Performance  on the validation data of \datafbn{}, \datawnn{}, and \datayago{} with different numbers of feature permutations.  We find that although increasing the number of permutations improves performance, it saturates as we exceed a certain limit. Please see Section \ref{sec:perm_kernel_results} for details.}
\end{figure}

\begin{table*}[t]
	\centering
\begin{tabular}{lm{3em}ccccccccccc}
	\toprule
	
	{} &  {} & \multicolumn{3}{c}{\textbf{RotatE}} && \multicolumn{3}{c}{\textbf{ConvE}} && \multicolumn{3}{c}{\textbf{\method{}}} \\
	\cmidrule(r){3-5}  \cmidrule(r){7-9} \cmidrule(r){11-13}
	{} & &   MRR & MR   & H@10  &&  MRR & MR   & H@10  &&  MRR & MR   & H@10   \\
	\midrule
		\multirow{4}{*}{\rotatebox[origin=c]{90}{Head Pred}}& 1-1 & \textbf{0.498}	& 359	& \textbf{0.593	}&& 0.374	& 223	& 0.505	&& 0.386	& \textbf{175}	& 0.547 \\
		& 1-N & 0.092	& 614	& 0.174	&& 0.091	& 700	& 0.17	&& \textbf{0.106}	& \textbf{573}	& \textbf{0.192} \\
		& N-1 & \textbf{0.471}	& 108	& \textbf{0.674}	&& 0.444	& 73	& 0.644	&& 0.466	& \textbf{69}	& 0.647 \\
		& N-N & 0.261	& \textbf{141}	& \textbf{0.476}	&& 0.261	& 158	& 0.459	&& \textbf{0.276}	& 148	& \textbf{0.476} \\
		\midrule
		\multirow{4}{*}{\rotatebox[origin=c]{90}{Tail Pred}}& 1-1 & \textbf{0.484}	& 307	& \textbf{0.578} && 0.366	& \textbf{261}	& 0.51	&& 0.368	& 308	& 0.547 \\
		& 1-N & 0.749	& 41	& 0.674	&& 0.762	& 33	& 0.878	&& \textbf{0.777}	& \textbf{27}	&\textbf{ 0.881} \\
		& N-1 & \textbf{0.074}	& \textbf{578}	& 0.138	&& 0.069	& 682	& 0.15	&& \textbf{0.074}	& 625	& \textbf{0.141} \\
		& N-N & 0.364	& \textbf{90}	& 0.608	&& 0.375	& 100	& 0.603	&& \textbf{0.395}	& 92	& \textbf{0.617} \\
	\bottomrule
\end{tabular}
	\caption{\label{tbl:results_rel_cat} Link prediction results by relation category on FB15k-237 dataset for RotatE, ConvE, and \method{}. Following (Wang et al., 2014b), the relations are categorized into one-to-one (1-1), one-to-many (1-N), many-to-one (N-1), and many-to-many (N-N). We observe that \method{} is effective at capturing complex relations compared to RotatE. Refer to Section \ref{sec:results_rel_cat} for details.}
\end{table*}

\subsection{Effect of  Feature Permutations}
\label{sec:perm_kernel_results}
In this section, we analyze the effect of increasing the number of feature permutations on \method{}'s performance on validation data of \datafbn{}, \datawnn{}, and \datayago{}. The overall results are summarized in Figure \ref{fig:perm_analysis}. 
We observe that on increasing the number of permuations although on \datafbn{}, MRR remains the same, it improves on \datawnn{} and \datayago{} datasets. However, it degrades as the number of permutations is increased beyond a certain limit. 
We hypothesize that this is due to over-parameteralization of the model. Moreover, since the number of relevant interactions are finite, increasing the number of permutations could become redundant beyond a limit.

\subsection{Evaluation on different Relation Types}
\label{sec:results_rel_cat}
In this section, we analyze the performance of \method{} on different relation categories of \datafbn{}. We chose \datafbn{} for analysis over other datasets because of its more and diverse set of relations. Following \cite{kg_relation_cat}, we classify the relations based on the average number of tails per head and heads per tail into four categories: one-to-one, one-to-many, many-to-one, and many-to-many. The results are presented in Table \ref{tbl:results_rel_cat}. Overall, we find that \method{} is effective at modeling complex relation types like one-to-many and many-to-many whereas, RotatE captures simple relations like one-to-one better. This demonstrates that an increase in interaction allows the model to capture more complex relationships.

\section{Conclusion}
\label{sec:conclusion}

In this paper, we propose \method{}, a novel knowledge graph embedding method which alleviates the limitations of ConvE by capturing additional heterogeneous feature interactions. \method{} is able to achieve this by utilizing three central ideas, namely feature permutation, checkered feature reshaping, and circular convolution. Through extensive experiments, we demonstrate that \method{} achieves a consistent improvement on link prediction performance on multiple datasets. We also theoretically analyze the effectiveness of the components of \method{}, and provide empirical validation of our hypothesis that increasing heterogeneous feature interaction is beneficial for link prediction with ConvE. This work demonstrates a possible scope for improving existing knowledge graph embedding methods by leveraging rich heterogenous interactions.


\bibliography{references.bib}
\bibliographystyle{aaai}

\appendix
	\section{Proof of Propositions}
	\addtocounter{section}{6}

	\begin{proposition}
		For any kernel $\bm{w}$ of size $k$, for all $n\ge\left(\frac{5k}{3}-1\right)$ if $k$ is odd and $n\ge\frac{(5k+2)(k-1)}{3k}$ if $k$ is even, the following statement holds:
		\[
		\mathcal{N}_{het}(\phi_{alt}, k) \ge \mathcal{N}_{het}(\phi_{stk}, k)
		\]
	\end{proposition}
	\begin{proof}
		\label{proof:thm1}
		Any $M_k \in \mathbb{R}^{k \times k}, \ M_k \subseteq \phi_{alt}$ contains $ \left\lfloor\frac{k}{2}\right\rfloor$ rows of elements of $\emb{s}$, and $\left\lfloor\frac{k+1}{2}\right\rfloor$ rows of elements of $\emb{r}$, or vice-versa.

		For a single fixed $M_k$, the total number of triples $(a_i, b_j, M_k)$ and $(b_j, a_i, M_k)$ is
		$$2 \times k \left\lfloor\dfrac{k}{2}\right\rfloor \times k \left\lfloor\dfrac{k+1}{2}\right\rfloor$$

		The number of possible $M_k$ matrices is $(n-k+1)^2$. Hence the total number of heterogeneous interactions is
		\begin{align*}
		\mathcal{N}_{het}(\phi_{alt}, k) = (n-k+1)^2 k^2 \times 2
		\left\lfloor\dfrac{k}{2}\right\rfloor \left\lfloor\dfrac{k+1}{2}\right\rfloor
		\end{align*}

		Any $M_k \in \mathbb{R}^{k \times k}, \ M_k \subseteq \phi_{stk}$ contains $l$ rows of elements of $\emb{s}$, and $k-l$ rows of elements of $\emb{r}$, where $0 \le l \le k$.

		For a fixed $l$, the number of different possible $M_k$ matrices is $(n-k+1)$. Hence, the number of heterogeneous interactions is
		\begingroup\makeatletter\def\f@size{9.5}\check@mathfonts
		\begin{align}
		& \mathcal{N}_{het}(\phi_{stk}, k)
		= (n-k+1)  \left( \sum_{l = 0}^{k} 2 \times kl \times k(k-l)\right) \nonumber \\
		&= (n-k+1) \cdot k^2 \cdot \left( k^2 (k + 1) - \dfrac{k(k+1)(2k+1)}{3} \right) \nonumber \\
		&= (n-k+1)\cdot k^2 \cdot \left( \dfrac{k(k+1)(k-1)}{3}\right) \label{eqn:hetcat}
		\end{align}
		\endgroup

		\noindent We need to check whether,
		$$\mathcal{N}_{het}(\phi_{alt}, k) \ge \mathcal{N}_{het}(\phi_{stk}, k)$$
		\begingroup\makeatletter\def\f@size{10}\check@mathfonts
		$$\implies (n-k+1) \left\lfloor\dfrac{k}{2}\right\rfloor \left\lfloor\dfrac{k+1}{2}\right\rfloor \ge \dfrac{k(k+1)(k-1)}{6}$$
		\endgroup

		\noindent For odd $k$, this becomes
		\begingroup\makeatletter\def\f@size{10}\check@mathfonts
		\begin{align*}
		(n-k+1)\left(\dfrac{k-1}{2}\right)\left(\dfrac{k+1}{2}\right) &\ge \dfrac{k(k+1)(k-1)}{6} \\
		n - k + 1 &\ge \dfrac{2k}{3} \\
		n &\ge \dfrac{5k}{3} - 1
		\end{align*}
		\endgroup

		\noindent For even $k$,
		\begin{align*}
		(n-k+1)\left(\dfrac{k}{2}\right)^2 &\ge \dfrac{k(k+1)(k-1)}{6} \\
		nk - k(k-1) &\ge \dfrac{2(k+1)(k-1)}{3} \\
		n &\ge \dfrac{(5k+2)(k-1)}{3k}
		\end{align*}

	\end{proof}

	\begin{proposition}
		For any kernel $\bm{w}$ of size $k$ and for all $\tau < \tau'$ ($\tau, \tau' \in \mathbb{N}$), the following statement holds:
		\[
		\mathcal{N}_{het}(\phi_{alt}^{\tau}, k) \ge \mathcal{N}_{het}(\phi_{alt}^{\tau'}, k)
		\]
	\end{proposition}
	\begin{proof}
		For simplicity, let us assume that $\alpha = n/(2\tau) \in \mathbb{N}$, i.e., $\phi_{alt}^\tau$ is composed of exactly $\alpha$ blocks of $\tau$ rows of $\emb{s}$ and $\emb{r}$ stacked alternately. Also, when $\tau < k$, we assume that $k/\tau \in \mathbb{N}$. Now, for any $M_k \in \mathbb{R}^{k \times k}, \ M_k \subseteq \phi_{alt}^\tau$, we consider the following two cases:
		\begin{case}
			$\bm{\tau \geq k-1}$:
			It is easy to see that this case can be split into $n/\tau$ subproblems, each of which is similar to $\phi_{stk}$. Hence,
			\[
				\m{N}_{het}(\phi_{alt}^\tau, k) = \left(\dfrac{n}{\tau}\right) \m{N}_{het}(\phi_{stk}, k)
			\]
		Clearly, $\m{N}_{het}(\phi_{alt}^\tau, k)$ is monotonically decreasing with increasing $\tau$.
		\end{case}
		\begin{figure}
			\centering
			\includegraphics[width=0.5\columnwidth]{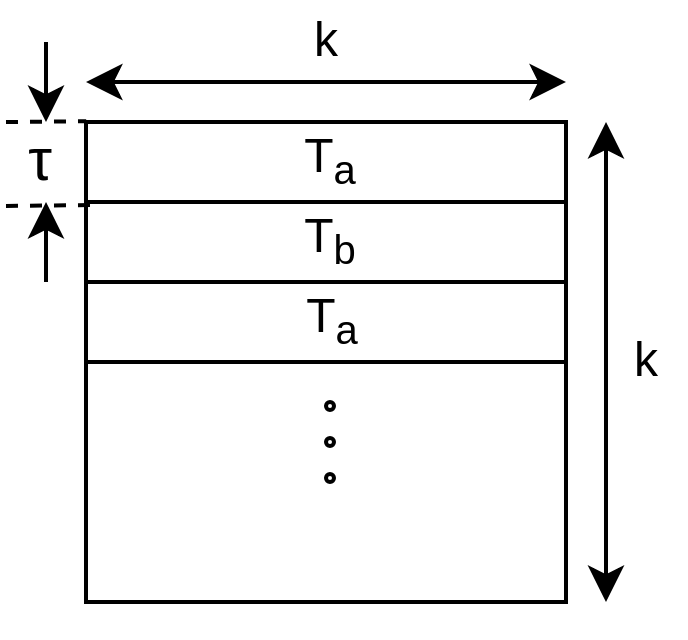}
			\caption{\label{fig: thm2} The figure depicts a $k \times k$ matrix $M_k$. $T_a, T_b$ are reshaped matrices each containing $\tau k$ components of $\emb{s}, \emb{r}$ respectively.}
		\end{figure}
		\begin{case}
			$\bm{\tau < k-1}$:
			As shown in Fig. \ref{fig: thm2}, let $T_a$, $T_b \in \mathbb{R}^{\tau \times k} $ denote a submatrix formed by components of $\emb{s}$, $\emb{r}$ respectively. Note that if $k$ is even, then for any $M_k \subseteq \phi_{alt}^\tau$, the number of components of $\emb{s}$ and $\emb{r}$ are always equal to  $k^2/2$ each. For odd $k$, the number of $T_a$'s and $T_b$'s are $\frac{(k/\tau)+1}{2}$ and $\frac{(k/\tau)-1}{2}$ in some order. Now, if we move $M_k$ down by $i$ rows ($i \leq \tau$), the total number of heterogeneous interactions across all such positions is:
			\begingroup\makeatletter\def\f@size{9}\check@mathfonts
				\begin{align*}
				& \dfrac{n}{\tau} \sum_{i=0}^{\tau-1} k^2\left( \left( \dfrac{k/\tau+1}{2}\right) \tau - i \right)  \left( \left( \dfrac{k/\tau-1}{2}\right) \tau +  i \right) \\
				&= \dfrac{nk^2}{\tau} \sum_{i=0}^{\tau-1} \dfrac{(k+\tau-2i)(k-\tau+2i)}{4} \\
				&= \dfrac{nk^2}{4\tau} \sum_{i=0}^{\tau-1} k^2 -(\tau^2 +4i^2 -4i\tau) \\
				&= \dfrac{nk^2}{4} \left( (k^2 - \tau^2) - \dfrac{4(\tau-1)(2\tau-1)}{6} + \dfrac{4\tau(\tau-1)}{2}\right)  \\
				&= C \left( k^2 - \dfrac{\tau^2}{3} - \dfrac{2}{3}\right)\\
				\end{align*}
			\endgroup
			We can see that this is also monotonically decreasing with increasing $\tau$. It is also evident that the above expression is maximum at $\tau = 1$ (since $\tau \in \mathbb{N}$).
		\end{case}
	\end{proof}

	\begin{proposition}
		For any kernel $\bm{w}$ of size $k$ and for all reshaping functions $\phi : \mathbb{R}^d \times \mathbb{R}^d \to \mathbb{R}^{n \times n}$, the following statement holds:
		\[
		\mathcal{N}_{het}(\phi_{chk}, k) \ge \mathcal{N}_{het}(\phi, k)
		\]
	\end{proposition}
	\begin{proof}
		For any $\phi$, and for any $M_{k} \in \mathbb{R}^{k \times k}$ such that $M_k \subseteq \phi$, let $x, y$ be the number of components of $\emb{s}$ and $\emb{r}$ in $M_k$ respectively. Then $\m{N}_{het}(M_k, k) = 2xy$. Also, since total number of elements in $M_k$ is fixed, we have $x + y = k^2$.

		Using the AM-GM inequality on $x, y$ we have,
		\[
		xy \leq \left(\dfrac{x+y}{2}\right)^2 = \dfrac{k^4}{4}
		\]
		If $k$ is odd, since $x,y \in \mathbb{N}$,
		\[
		xy \leq \dfrac{k^4 - 1}{4}
		\]
		Therefore, the maximum interaction occurs when $x = y = \frac{k^2}{2}$ (for even $k$), or $x = \frac{k^2+1}{2},y = \frac{k^2 - 1}{2}$ (for odd $k$). It can be easily verified that this property holds for all $M_{k} \subseteq \phi_{chk}$. Hence,

		\[
		\m{N}_{het}(\phi, k) = \sum_{M_k} 2xy \le \sum_{M_k} \dfrac{2k^4}{4} = \m{N}_{het}(\phi_{chk}, k)
		\]
	\end{proof}

	\begin{proof}
		If $M_k$ contains $x$ components of $\emb{s}$ and $y$ components of $\emb{r}$, then $\m{N}_{het}(M_k, k) = 2xy$, and $\m{N}_{het}(M'_k, k) = 2(x-l)(y-(p-l))$ for some $l \le p$ and $l \le x$. We observe that
		\begingroup\makeatletter\def\f@size{10}\check@mathfonts
		\begin{align*}
		\m{N}_{het}(M'_k, k) &= \m{N}_{het}(M_k, k)-2(x-l)(p-l) -2ly\\
		&\le \m{N}_{het}(M_k, k)
		\end{align*}
		\endgroup
	\end{proof}

	\begin{proposition}
		Let $\Omega_0$, $\Omega_c : \mathbb{R}^{n \times n} \to \mathbb{R}^{(n+p) \times (n+p)}$ denote zero padding and circular padding functions respectively, for some $p > 0$. Then for any reshaping function $\phi$,
		\[
		\mathcal{N}_{het}(\Omega_c(\phi), k) \ge \mathcal{N}_{het}(\Omega_0(\phi), k)
		\]
	\end{proposition}
	\begin{proof}
		Given $\Omega_c(\phi))$, we know that we can obtain $\Omega_0(\phi)$ by replacing certain components of $\Omega_c(\phi))$ with $0$. So for every $M_k \subseteq \Omega_c(\phi)$, there is a corresponding $M'_k \subseteq \Omega_0(\phi)$ which is obtained by replacing some $p$ components ($p \ge 0$) of $M_k$ with $0$.

		Using the above Lemma, we can see that
		\begin{align*}
		\m{N}_{het}(\Omega_c(\phi), k)) &= \sum_{M_k \subseteq \Omega_c(\phi)} \m{N}_{het}(M_k, k)) \\
		&\ge \sum_{M'_k \subseteq \Omega_0(\phi)} \m{N}_{het}(M'_k, k)) \\
		&= \m{N}_{het}(\Omega_0(\phi), k))
		\end{align*}

	\end{proof}

\setcounter{section}{1}
\section{Hyperparameters}
\label{sec:hyperparams}
We use the standard training, validation and test splits provided with the datasets. A detailed description of the datasets is included in the main paper. We select the best model using the validation data on the hyperparameters listed in Table \ref{table:hyperparams}. Most of the hyperparamters are adopted from ConvE \cite{conve} model. In this paper, we explore both 1-1 \cite{transe} and 1-N \cite{conve} scoring
techniques. In 1-N scoring, each $(s, r)$ pair is scored against
all the entities $o \in \m{E}$ simultaneously.
For training, we use Adam optimizer \cite{adam_opt} and Xavier initialization \cite{xavier_init} for initializing parameters.
	
\begin{table}[h]
	\centering
	\small
	\begin{tabular}{ll}
		\toprule
		Hyperparameter                 & Values 					\\
		\midrule
		Learning rate                  & \{0.001, 0.0001\}			\\
		Label smoothening              & \{0.0, 0.1\}			\\
		Batch size                     & \{16, 64, 256\}			\\
		Negative Samples                     & \{100, 500, 1000, 4000\}			\\
		$l_2$ regularization		   & \{0, $10^{-5}$\}			\\
		Hidden dropout                 & \{0, 0.3, 0.5\}			\\
		Feature dropout                & \{0, 0.2, 0.5\}			\\
		Input dropout                  & \{0, 0.2\}			\\
		$k_w$                          & \{10\}					\\
		$k_h$                          & \{20\}					\\
		Number of convolutional filters& \{32, 48, 64, 96\}			\\
		Convolutional kernal size $k$  & \{3, 5, 7, 9, 11\}		\\
		Number of feature permutations $t$   & \{1, 2, 3, 4, 5\}				\\
		\bottomrule
	\end{tabular}
	\caption{Details of hyperparameters used. Please refer to Section \ref{sec:hyperparams} for more details.}
	\label{table:hyperparams}
\end{table}

\end{document}